\documentclass[10pt,twocolumn,letterpaper]{article}

\usepackage{cvpr}
\usepackage{amsmath}
\usepackage{amssymb}
\usepackage{siunitx}
\usepackage{booktabs}
\usepackage{array}
\usepackage{multirow}
\usepackage{times}
\usepackage{graphicx}
\usepackage{subfigure}

\usepackage{enumitem}
\setenumerate[1]{itemsep=0pt,partopsep=0pt,parsep=\parskip,topsep=5pt}
\setitemize[1]{itemsep=0pt,partopsep=0pt,parsep=\parskip,topsep=5pt}
\setdescription{itemsep=0pt,partopsep=0pt,parsep=\parskip,topsep=5pt}

\usepackage[breaklinks=true,bookmarks=false]{hyperref}

\cvprfinalcopy 

\def\cvprPaperID{7288}

\ifcvprfinal\fi
\begin{document}

\title{GraphTER: Unsupervised Learning of Graph Transformation Equivariant Representations via Auto-Encoding Node-wise Transformations}

\author{Xiang Gao\textsuperscript{1}, 
Wei Hu\textsuperscript{1,}\thanks{Corresponding author: Wei Hu (forhuwei@pku.edu.cn). 
This work was supported by National Natural Science Foundation of China [61972009] and Beijing Natural Science Foundation [4194080].}\;,
and Guo-Jun Qi\textsuperscript{2}\\
\textsuperscript{1}Wangxuan Institute of Computer Technology, Peking University, Beijing\\
\textsuperscript{2}Futurewei Technologies\\
{\tt\small \{gyshgx868, forhuwei\}@pku.edu.cn, guojunq@gmail.com}
}

\maketitle
\thispagestyle{empty}

\begin{abstract}
Recent advances in Graph Convolutional Neural Networks (GCNNs) have shown their efficiency for non-Euclidean data on graphs, which often require a large amount of labeled data with high cost. 
It it thus critical to learn graph feature representations in an unsupervised manner in practice. 
To this end, we propose a novel unsupervised learning of Graph Transformation Equivariant Representations (GraphTER), aiming to capture intrinsic patterns of graph structure under both global and local transformations. 
Specifically, we allow to sample different groups of nodes from a graph and then transform them node-wise isotropically or anisotropically. 
Then, we self-train a representation encoder to capture the graph structures by reconstructing these node-wise transformations from the feature representations of the original and transformed graphs. 
In experiments, we apply the learned GraphTER to graphs of 3D point cloud data, and results on point cloud segmentation/classification show that GraphTER significantly outperforms state-of-the-art unsupervised approaches and pushes greatly closer towards the upper bound set by the fully supervised counterparts.
The code is available at: \href{https://github.com/gyshgx868/graph-ter}{https://github.com/gyshgx868/graph-ter}.
\end{abstract}

\vspace{-0.1in}
\section{Introduction}
\vspace{-0.05in}
\label{sec:intro}

Graphs are a natural representation of  irregular data, such as 3D geometric points, social networks, citation networks and brain networks. 
Recent advances in Graph Convolutional Neural Networks (GCNNs) have shown their efficiency in learning representations of such data \cite{bronstein2017geometric,zhou2018graph,zhang2018deep,wu2019comprehensive}, which generalize the celebrated CNN models. 
Existing GCNNs are mostly trained in a supervised or semi-supervised fashion, requiring a large amount of labeled data to learn graph representations.
This prevents the wide applicability of GCNNs due to the high labeling cost especially for large-scale graphs in many real scenarios. 
Hence, it is demanded to train graph feature representations in an unsupervised fashion, which can adapt to downstream learning tasks on graphs. 

\begin{figure}[t]
    \centering
    \includegraphics[width=8.3cm]{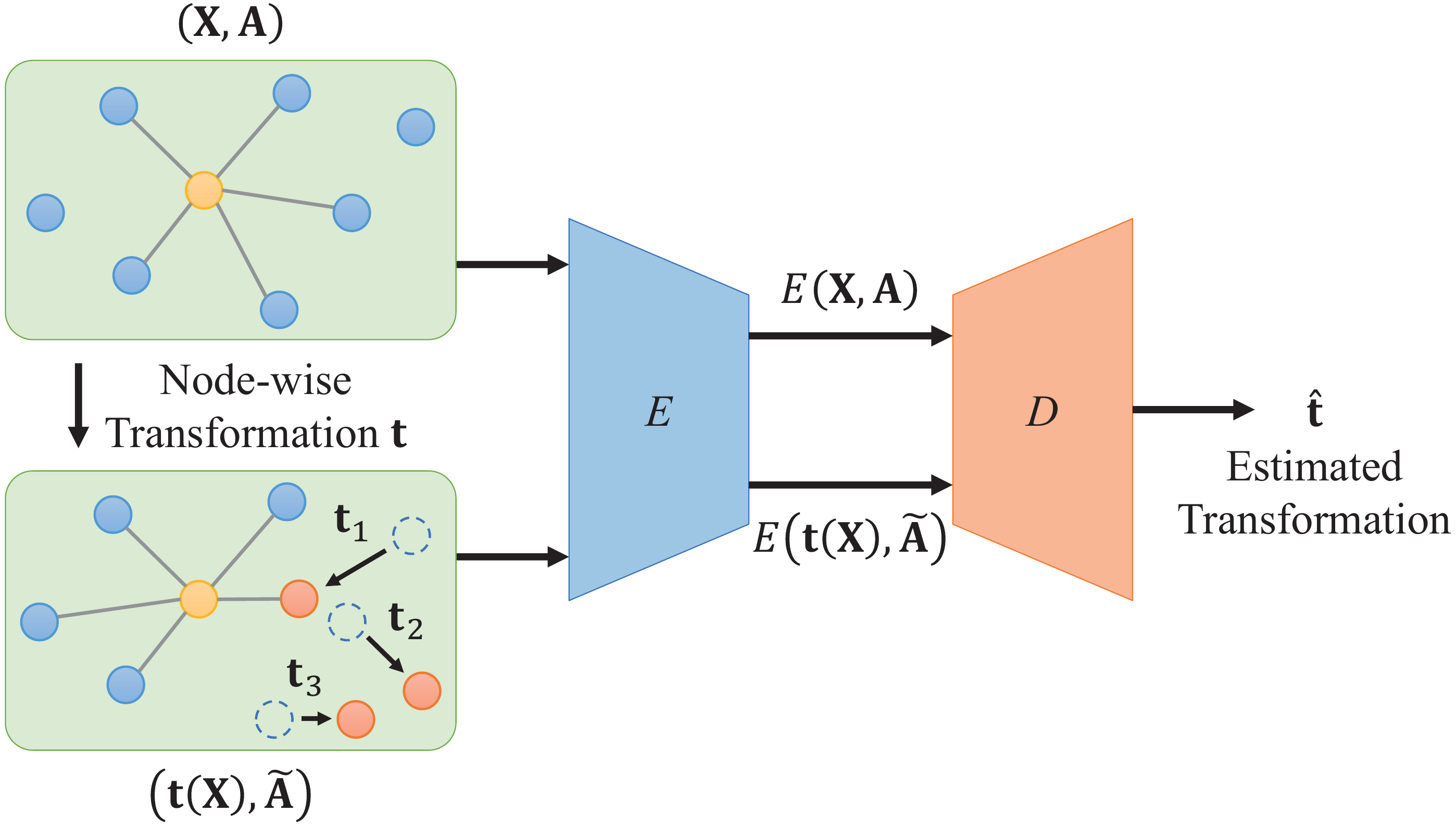}
    \caption{\textbf{An illustration of the proposed GraphTER model for unsupervised feature learning.}
    The encoder learns representations of the original graph data $\mathbf{X}$ associated with adjacency matrix $\mathbf{A}$ and its transformed counterpart $\mathbf{t}(\mathbf{X})$ associated with $\tilde{\mathbf{A}}$ respectively. 
    By decoding node-wise transformations $\mathbf t$ from both representations, the auto-encoder is able to learn intrinsically morphable structures of graphs.}
    \label{fig:teaser}
    \vspace{-0.2in}
\end{figure}

Auto-Encoders (AEs) and Generative Adversarial Networks (GANs) are two most representative methods for unsupervised learning. 
Auto-encoders aim to train an encoder to learn feature representations by reconstructing input data via a decoder \cite{vincent2008extracting,rifai2011contractive,hinton2011transforming,kingma2013auto}. 
Most of auto-encoders stick to the idea of reconstructing \textit{input data} at the output end ({\it e.g.}, images \cite{hinton2011transforming}, graphs \cite{kipf2016variational}, 3D point clouds \cite{yang2018foldingnet}), and thus can be classified into the Auto-Encoding Data (AED) \cite{zhang2019aet}.
In contrast, GANs \cite{goodfellow2014generative,donahue2016adversarial,dumoulin2016adversarially,you2018graphrnn,li2018learning,de2018molgan,bojchevski2018netgan} extract feature representations in an unsupervised fashion by generating data from input noises via a pair of generator and discriminator, where the noises are viewed as the data representations, and the generator is trained to generate data from the ``noise" feature representations. 

Based on AEs and GANs, many approaches have sought to learn \textit{transformation equivariant representations} (TER) to further improve the quality of unsupervised representation learning. It assumes that representations equivarying to transformations are able to encode the intrinsic structures of data such that the transformations can be reconstructed from the representations before and after transformations \cite{qi2019learning}. 
Learning transformation equivariant representations has been advocated in Hinton's seminal work on learning transformation capsules \cite{hinton2011transforming}.
Following this, a variety of approaches have been proposed to learn transformation equivariant representations \cite{kivinen2011transformation,sohn2012learning,schmidt2012learning,skibbe2013spherical,lenc2015understanding,gens2014deep,dieleman2015rotation,dieleman2016exploiting}.
Among them are the group equivariant convolutions \cite{cohen2016group} and group equivariant capsule networks \cite{lenssen2018group} that generalize the CNNs and capsule nets to equivary to various transformations.
However, these models are restricted to discrete transformations, and they should be trained in a supervised fashion. This limits their ability to learn unsupervised representations equivariant to a generic composition of continuous transformations \cite{qi2019learning}.

To generalize to generic transformations, Zhang {\it et al.} \cite{zhang2019aet} propose to learn unsupervised feature representations via Auto-Encoding Transformations (AET) rather than AED.
By randomly transforming images, they seek to train auto-encoders by directly reconstructing these transformations from the learned representations of both the original and transformed images.
A variational AET  \cite{qi2019avt} is also introduced from an information-theoretic perspective by maximizing the lower bound of mutual information between transformations and representations. Moreover, it has also been theoretically proved \cite{qi2019learning,qi2019avt} that the unsupervised representations by the AET are equivariant to the applied transformations. 
Unfortunately, these works focus on transformation equivariant representation learning of images that are Euclidean data, which cannot be extended to graphs due to the irregular data structures.

In this paper, we take a first step towards this goal --
we formalize Graph Transformation Equivariant Representation (GraphTER) learning by auto-encoding node-wise transformations in an unsupervised manner.
The proposed method is novel in twofold aspects. 
On one hand, we define \textit{graph signal transformations} and present a graph-based auto-encoder architecture, which encodes the representations of the original and transformed graphs so that the graph transformations can be reconstructed from both representations.
On the other hand, in contrast to the AET where global spatial transformations are applied to the \textit{entire} input image, we perform node-wise transformations on graphs, where each node can have its own transformation.
Representations of individual nodes are thus learned by decoding node-wise transformations to reveal the graph structures around it. 
These representations will not only capture the local graph structures under node-wise transformations, but also reveal global information about the graph as we randomly sample nodes into different groups over training iterations. Different groups of nodes model different parts of graphs, allowing the learned representations to capture  various scales of graph structures under isotropic and/or anisotropic node-wise transformations. This results in transformation equivariant representations to characterize the \textit{intrinsically} morphable structures of graphs.

Specifically, given an input graph signal and its associated graph, we sample a subset of nodes from the graph (globally or locally) and perform transformations on individual nodes, either isotropically or anisotropically.
Then we design a full graph-convolution auto-encoder architecture, where the encoder learns the representations of individual nodes in the original and transformed graphs respectively, and the decoder predicts the applied node-wise transformations from both representations.
Experimental results demonstrate that the learned GraphTER significantly outperforms the state-of-the-art unsupervised models, and achieves comparable results to the fully supervised approaches in the 3D point cloud classification and segmentation tasks.


Our main contributions are summarized as follows.
\begin{itemize}
    \item We are the first to propose Graph Transformation Equivariant Representation (GraphTER) learning to extract adequate graph signal feature representations in an unsupervised fashion.
    \item We define generic graph transformations and formalize the GraphTER to learn feature representations of graphs by decoding node-wise transformations end-to-end in a full graph-convolution auto-encoder architecture. 
    \item  Experiments demonstrate the GraphTER model outperforms the state-of-the-art methods in unsupervised graph feature learning as well as greatly pushes closer to the upper bounded performance by the fully supervised counterparts.
\end{itemize}

The remainder of this paper is organized as follows. 
We first review related works in Sec.~\ref{sec:related_works}. 
Then we define graph transformations in Sec.~\ref{sec:graphT} and formalize the GraphTER model in Sec.~\ref{sec:method}. 
Finally, experimental results and conclusions are presented in Sec.~\ref{sec:experiments} and Sec.~\ref{sec:conclusion}, respectively.

\vspace{-0.1in}
\section{Related Works}
\vspace{-0.05in}
\label{sec:related_works}
\textbf{Transformation Equivariant Representations.} Many approaches have been proposed to learn equivariant representations, including transforming auto-encoders \cite{hinton2011transforming}, equivariant Boltzmann machines \cite{kivinen2011transformation,sohn2012learning}, equivariant descriptors \cite{schmidt2012learning}, and equivariant filtering \cite{skibbe2013spherical}.
Lenc et al. \cite{lenc2015understanding} prove that the AlexNet \cite{krizhevsky2012imagenet} trained on ImageNet learns representations that are equivariant to flip, scaling and rotation transformations.
Gens {\it et al.} \cite{gens2014deep} propose an approximately equivariant convolutional architecture, which utilizes sparse and high-dimensional feature maps to deal with groups of transformations.
Dieleman {\it et al.} \cite{dieleman2015rotation} show that rotation symmetry can be exploited in convolutional networks for effectively learning an equivariant representation.
This work is later extended in \cite{dieleman2016exploiting} to evaluate on other computer vision tasks that have cyclic symmetry.
Cohen {\it et al.} \cite{cohen2016group} propose group equivariant convolutions that have been developed to equivary to more types of transformations. The idea of group equivariance has also been introduced to the capsule nets \cite{lenssen2018group} by ensuring the equivariance of output pose vectors to a group of transformations.
Zhang {\it et al.} \cite{zhang2019aet} propose to learn unsupervised feature representations via Auto-Encoding Transformations (AET) by estimating transformations from the learned feature representations of both the original and transformed images.
This work is later extended in \cite{qi2019avt} by introducing a variational transformation decoder, where the AET model is trained from an information-theoretic perspective by maximizing the lower bound of mutual information.

\textbf{Auto-Encoders and GANs.} Auto-encoders (AEs) have been widely adopted to learn unsupervised representations \cite{hinton1994autoencoders}, which employ an encoder to extract feature representations and a decoder to reconstruct the input data from the representations.
The idea is based on good feature representations should contain sufficient information to reconstruct the input data.
A large number of approaches have been proposed following this paradigm of Auto-Encoding Data (AED), including variational AEs (VAEs) \cite{kingma2013auto}, denoising AEs \cite{vincent2008extracting}, contrastive AEs \cite{rifai2011contractive}, transforming AEs \cite{hinton2011transforming}, {\it etc.}
Based on the above approaches, graph AEs have been proposed to learn latent representations for graphs.
These approaches basically learn graph embeddings for plain graphs \cite{cao2016deep,wang2016structural} and attributed graphs \cite{kipf2016variational,pan2018adversarially}, which are still trained in the AED fashion.
In addition to AEs, Generative Adversarial Networks (GANs) \cite{goodfellow2014generative} become popular for learning unsupervised representations of data, which tend to generate data from noises sampled from a random distribution.
The basic idea of these models is to treat the sampled noise as the feature of the output data, and an encoder can be trained to obtain the ``noise" feature representations for input data, while the generator is treated as the decoder to generate data from the ``noise"  feature representations \cite{donahue2016adversarial,dumoulin2016adversarially}.
Recently, several approaches have been proposed to build graph GANs. 
For instance, \cite{you2018graphrnn} and \cite{li2018learning} propose to generate nodes and edges alternately, while \cite{de2018molgan} and \cite{bojchevski2018netgan} propose to integrate GCNNs with LSTMs and GANs respectively to generate graphs.

\vspace{-0.1in}
\section{Graph Transformations}
\label{sec:graphT}
\vspace{-0.05in}
\subsection{Preliminaries}

Consider an undirected graph $\mathcal{G}=\{\mathcal{V},\mathcal{E}\}$ composed of a node set $\mathcal{V}$ of cardinality $|\mathcal{V}|=N$, and an edge set $\mathcal{E}$ connecting nodes. 
\textit{Graph signal} refers to data/features associated with the nodes of $\mathcal{G}$, denoted by $\mathbf{X} \in \mathbb{R}^{N \times C}$ with $i$th row representing the $C$-dimensional graph signal at the $i$th node of $\mathcal{V}$.

To characterize the similarities (and thus the graph structure) among node signals, an \textit{adjacency matrix} $\mathbf{A}$ is defined on $\mathcal{G}$, which is a real-valued symmetric $N \times N$ matrix with $a_{i,j}$ as the weight assigned to the edge $(i,j)$ connecting nodes $i$ and $j$. 
Formally, the adjacency matrix is constructed from graph signals as follows,
\begin{equation}
    \setlength{\abovedisplayskip}{3pt}
    \setlength{\belowdisplayskip}{3pt}
    \mathbf{A} = f(\mathbf{X}),
    \label{eq:A_X}
\end{equation}
where $f(\cdot)$ is a linear or non-linear function applied to each pair of nodes to get the pair-wise similarity. For example, a widely adopted function is to nonlinearly construct a $k$-nearest-neighbor ($k$NN) graph from node features \cite{wang2019dynamic,zhang2018graph}.

\begin{figure*}[t]
    \centering
    \subfigure[Original model]{
    \includegraphics[width=0.18\textwidth]{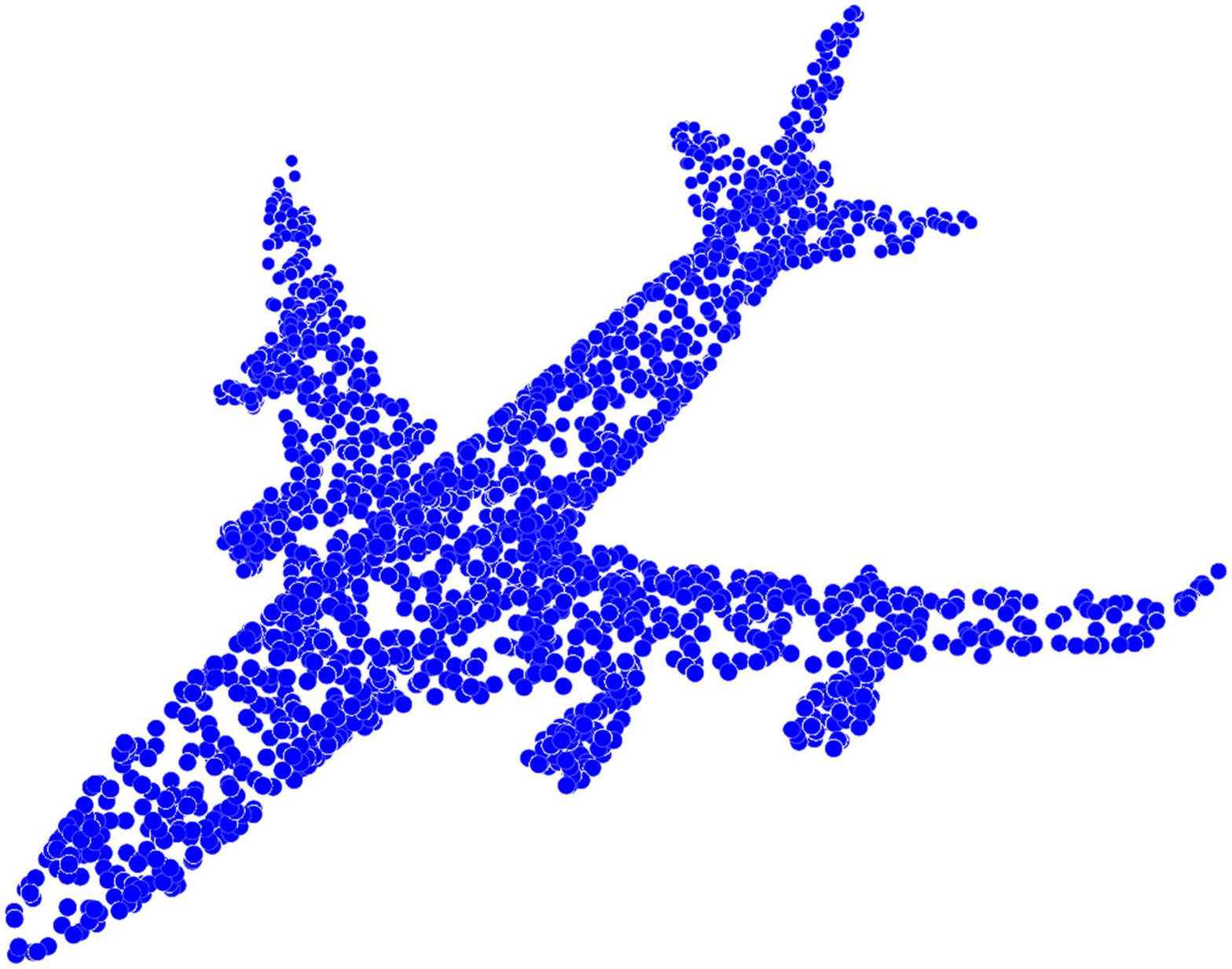}
    }
    \subfigure[Global+Isotropic]{
    \includegraphics[width=0.18\textwidth]{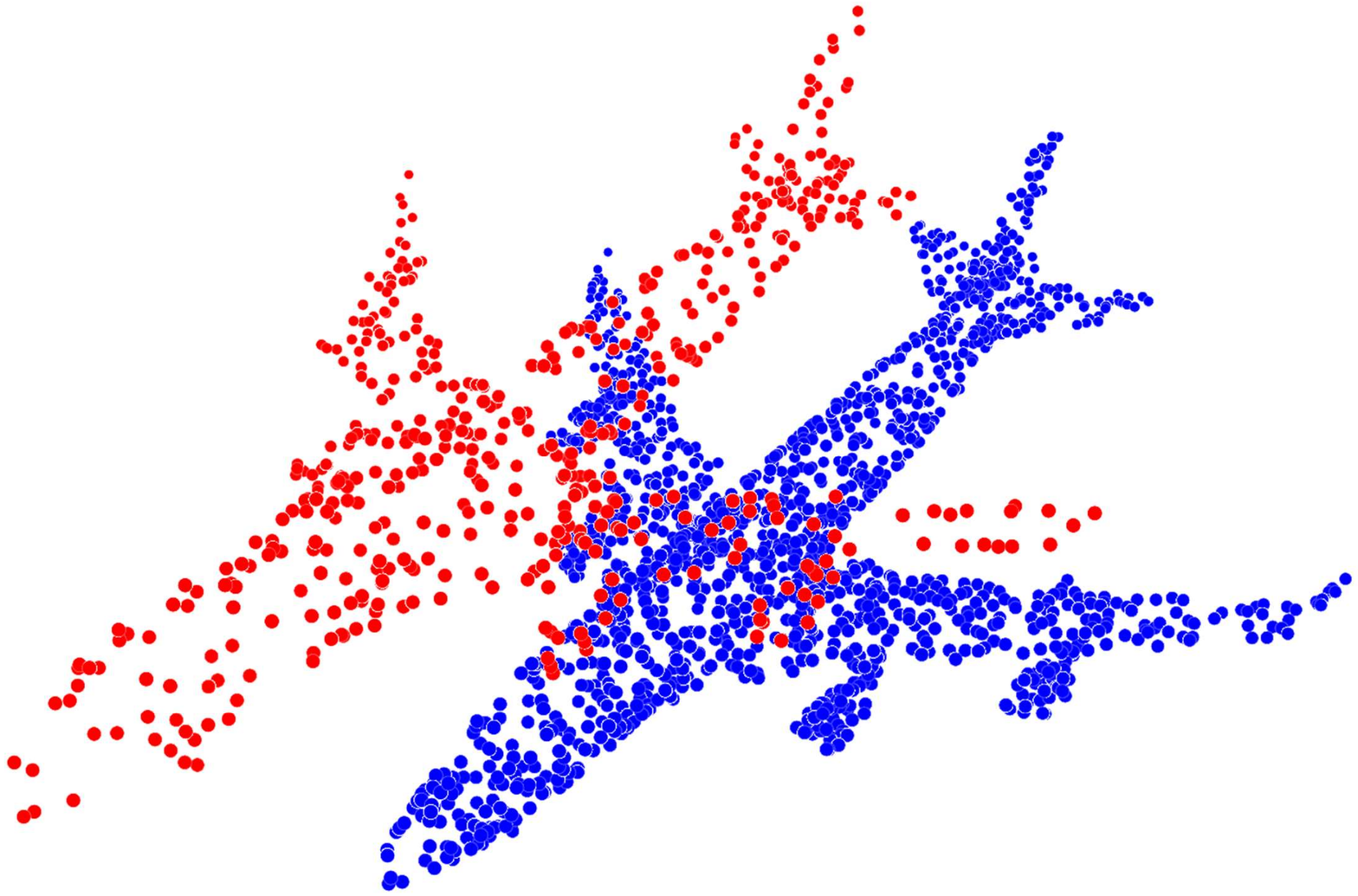}
    }
    \subfigure[Global+Anisotropic]{
    \includegraphics[width=0.18\textwidth]{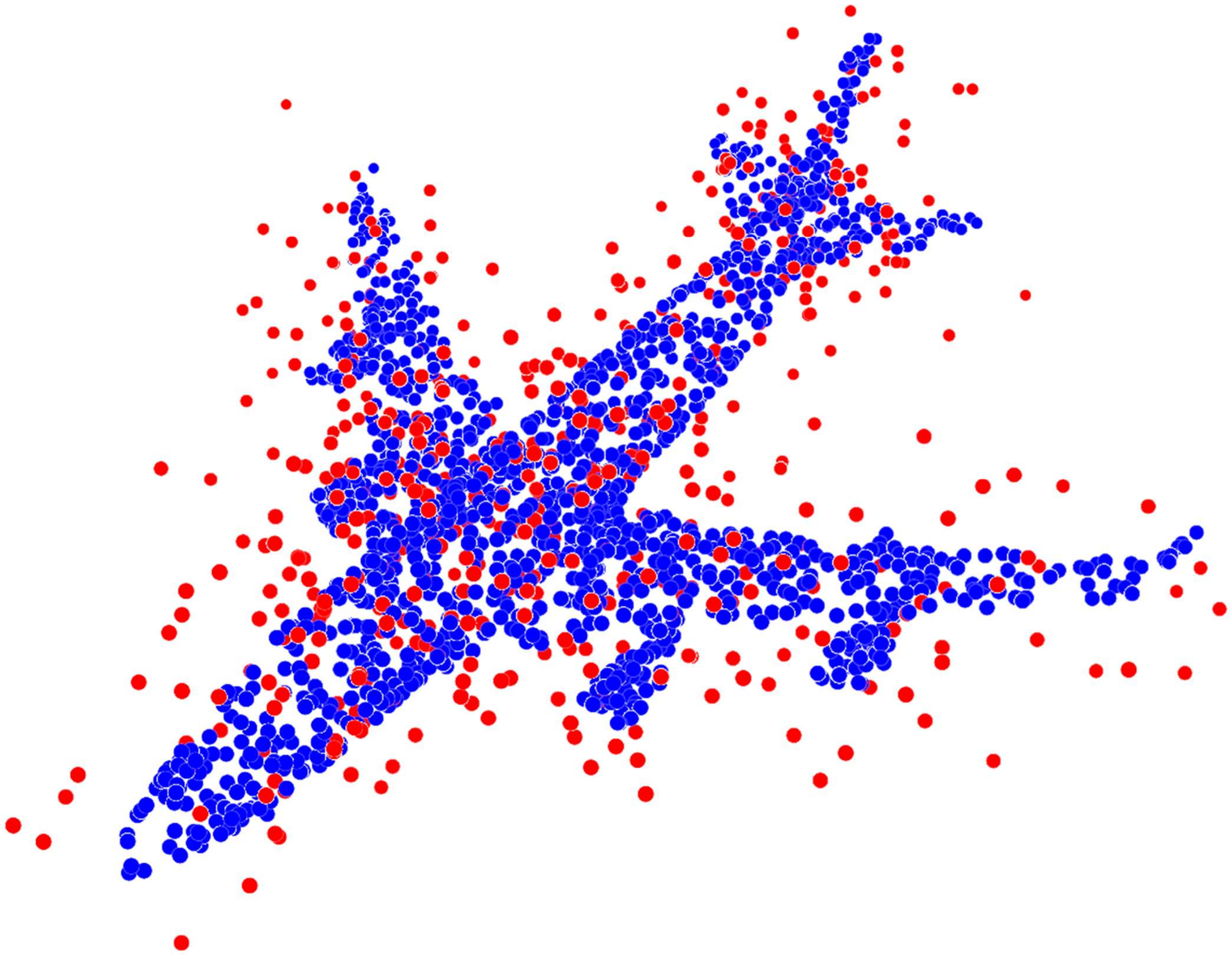}
    }
    \subfigure[Local+Isotropic]{
    \includegraphics[width=0.18\textwidth]{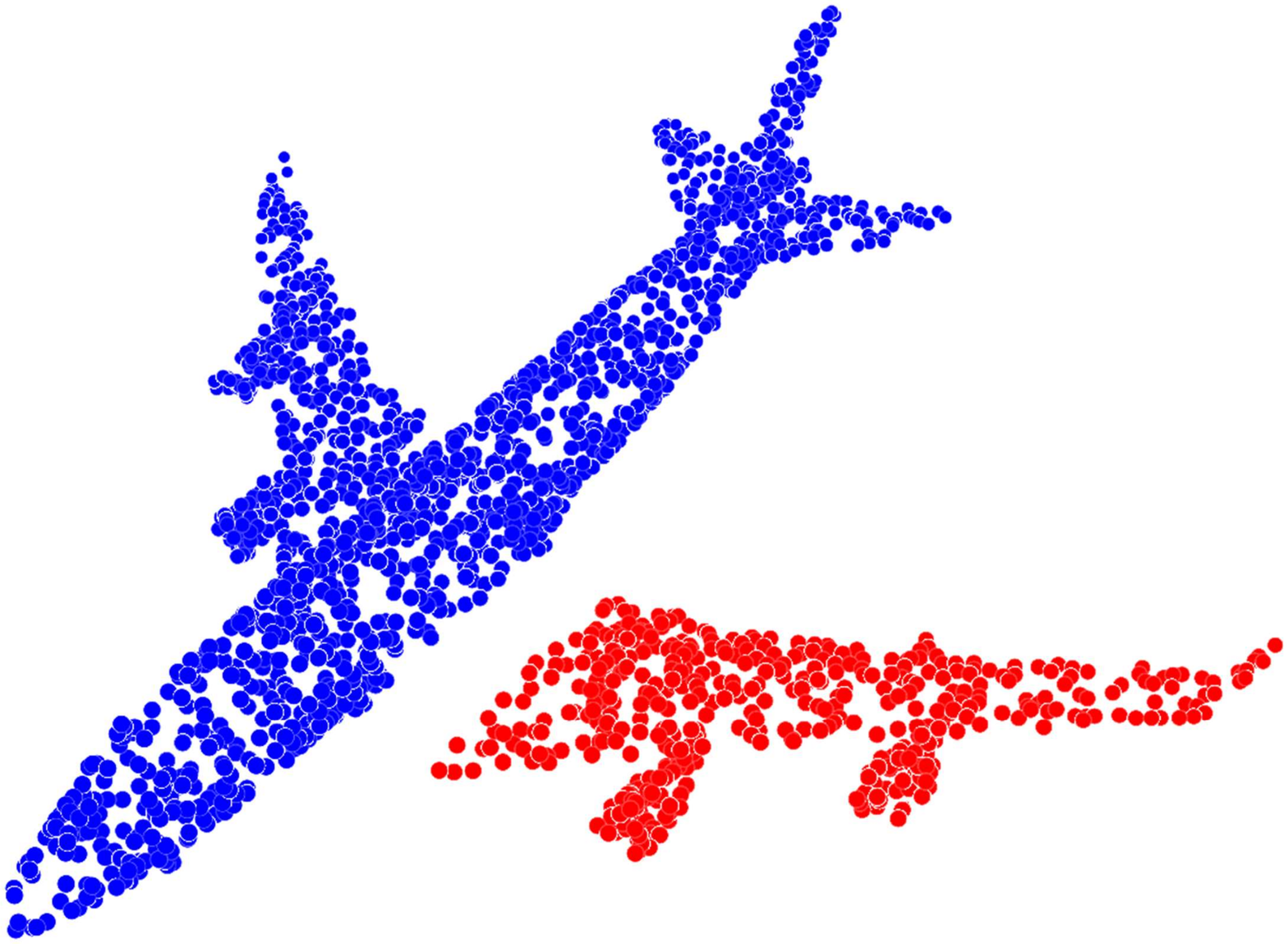}
    }
    \subfigure[Local+Anisotropic]{
    \includegraphics[width=0.18\textwidth]{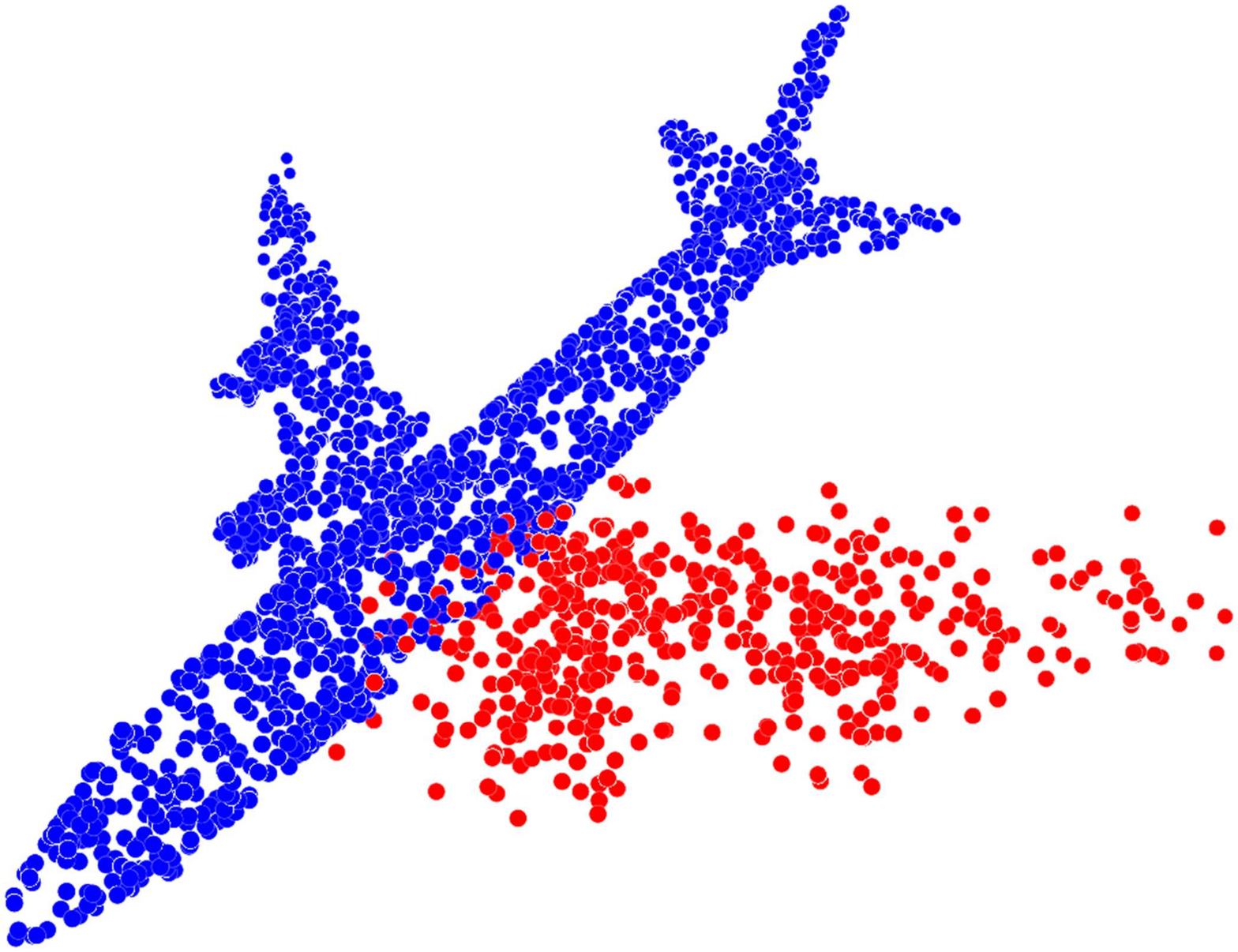}
    }
    \caption{\textbf{Demonstration of different sampling (Global or Local) and node-wise \textit{translation} (Isotropic or Anisotropic) methods on 3D point clouds.} Red and blue points represent transformed and non-transformed points, respectively. Note that we adopt the wing as a sampled local point set for clear visualization.}
    \label{fig:sampling_transform}
    \vspace{-0.2in}
\end{figure*}

\vspace{-0.05in}
\subsection{Graph Signal Transformation}

Unlike Euclidean data like images, graph signals are irregularly sampled, whose transformations are thus nontrivial to define.
To this end, we define a graph transformation on the signals $\mathbf{X}$ as node-wise \textit{filtering} on $\mathbf{X}$.

Formally, suppose we sample a graph transformation $\mathbf{t}$ from a transformation distribution $\mathcal{T}_g$, \textit{i.e.}, $\mathbf{t} \sim \mathcal{T}_g$. 
Applying the transformation to graph signals $\mathbf{X}$ that are sampled from data distribution $\mathcal{X}_g$, \textit{i.e.}, $\mathbf{X} \sim \mathcal{X}_g$, leads to the filtered graph signal
\begin{equation}
    \setlength{\abovedisplayskip}{3pt}
    \setlength{\belowdisplayskip}{3pt}
    \tilde{\mathbf{X}}=\mathbf{t}(\mathbf{X}).
    \label{eq:graph_signal_t}
\end{equation}

The filter $\mathbf t$ is applied to each node individually, which can be either node-invariant or node-variant. In other words, the transformation of each node signal associated with $\mathbf t$  can be different from each other. For example, for a  translation $\mathbf t$, a distinctive translation can be applied to each node. We will call the graph transformation isotropic (anisotropic) if it is node-invariant (variant).

Consequently, the adjacency matrix of the transformed graph signal $\tilde{\mathbf{X}}$ equivaries according to \eqref{eq:A_X}:
\begin{equation}
    \setlength{\abovedisplayskip}{3pt}
    \setlength{\belowdisplayskip}{3pt}
    \tilde{\mathbf{A}} = f(\tilde{\mathbf{X}}) = f(\mathbf{t}(\mathbf{X})),
\end{equation}
which transforms the \textit{graph structures}, as edge weights are also filtered by $\mathbf{t}(\cdot)$.

Under this definition, there exist a wide spectrum of graph signal transformations. 
Examples include affine transformations (translation, rotation and shearing) on the location of nodes (\textit{e.g.}, 3D coordinates in point clouds), and graph filters such as low-pass filtering on graph signals by the adjacency matrix \cite{sandryhaila2013discrete}.  

\vspace{-0.05in}
\subsection{Node-wise Graph Signal Transformation}

As aforementioned, in this paper, we focus on \textit{node-wise} graph signal transformation, {\it i.e.}, each node has its own transformation, either isotropically or anisotropically. 
We seek to learn graph representations through the node-wise transformations by revealing how different parts of graph structures would change globally and locally.

Specifically, here are two distinct advantages.
\begin{itemize}
    \item The node-wise transformations allow us to use node sampling to study different parts of graphs under various transformations. 
    \item By decoding the node-wise transformations, we will be able to learn the representations of individual nodes. Moreover, these node-wise representations will not only capture the local graph structures under these transformations, but also contain global information about the graph when these nodes are sampled into different groups over iterations during training.
\end{itemize}

Next, we discuss the formulation of learning graph transformation equivariant representations by decoding the node-wise transformations via a graph-convolutional encoder and decoder.




\vspace{-0.1in}
\section{GraphTER: The Proposed Approach}
\label{sec:method}
\begin{figure}[t]
    \centering
    \subfigure[Before transformation.]{
    \includegraphics[width=0.45\columnwidth]{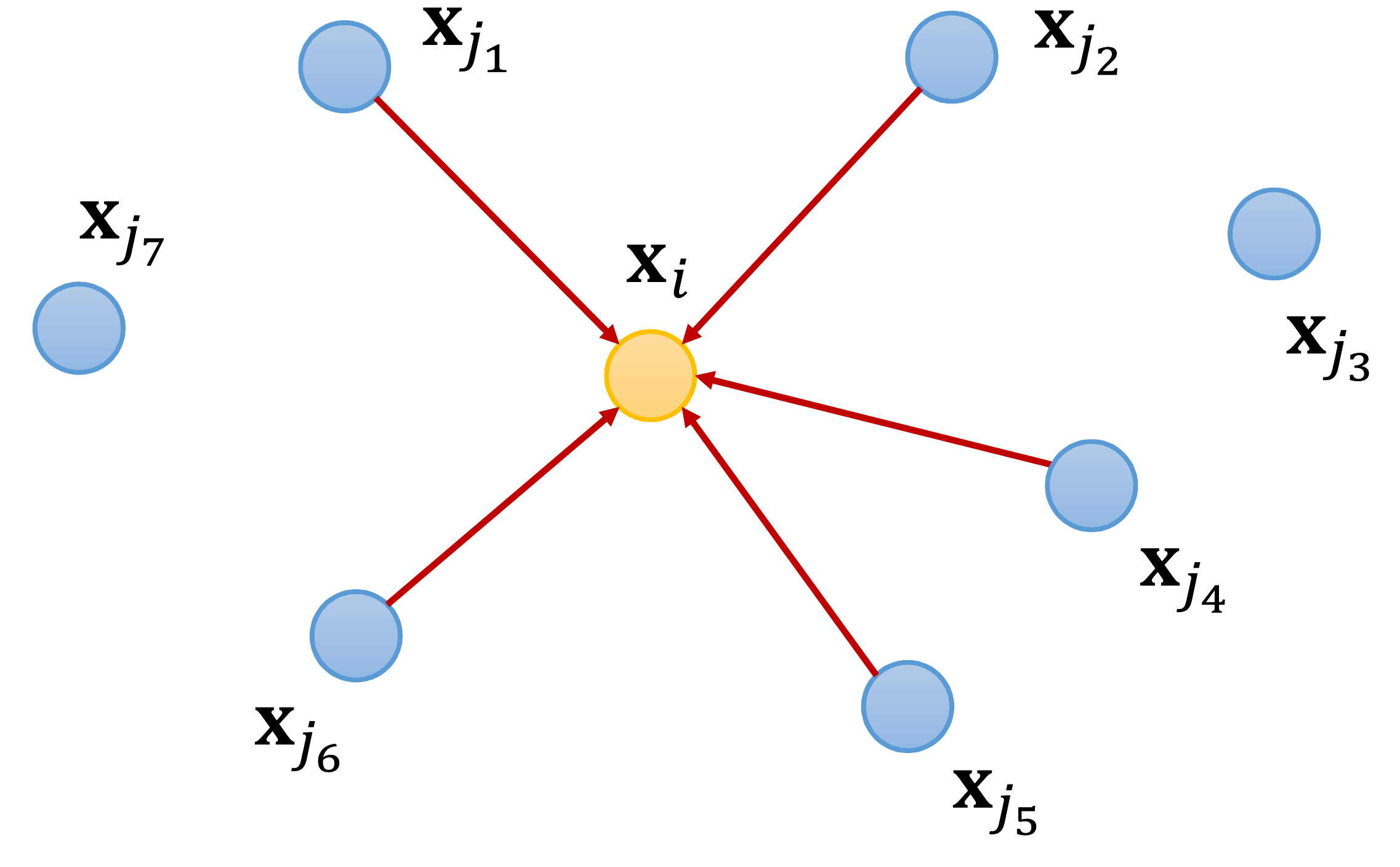}
    \label{subfig:before_transform}
    }
 	\subfigure[After transformation.]{
 	\includegraphics[width=0.45\columnwidth]{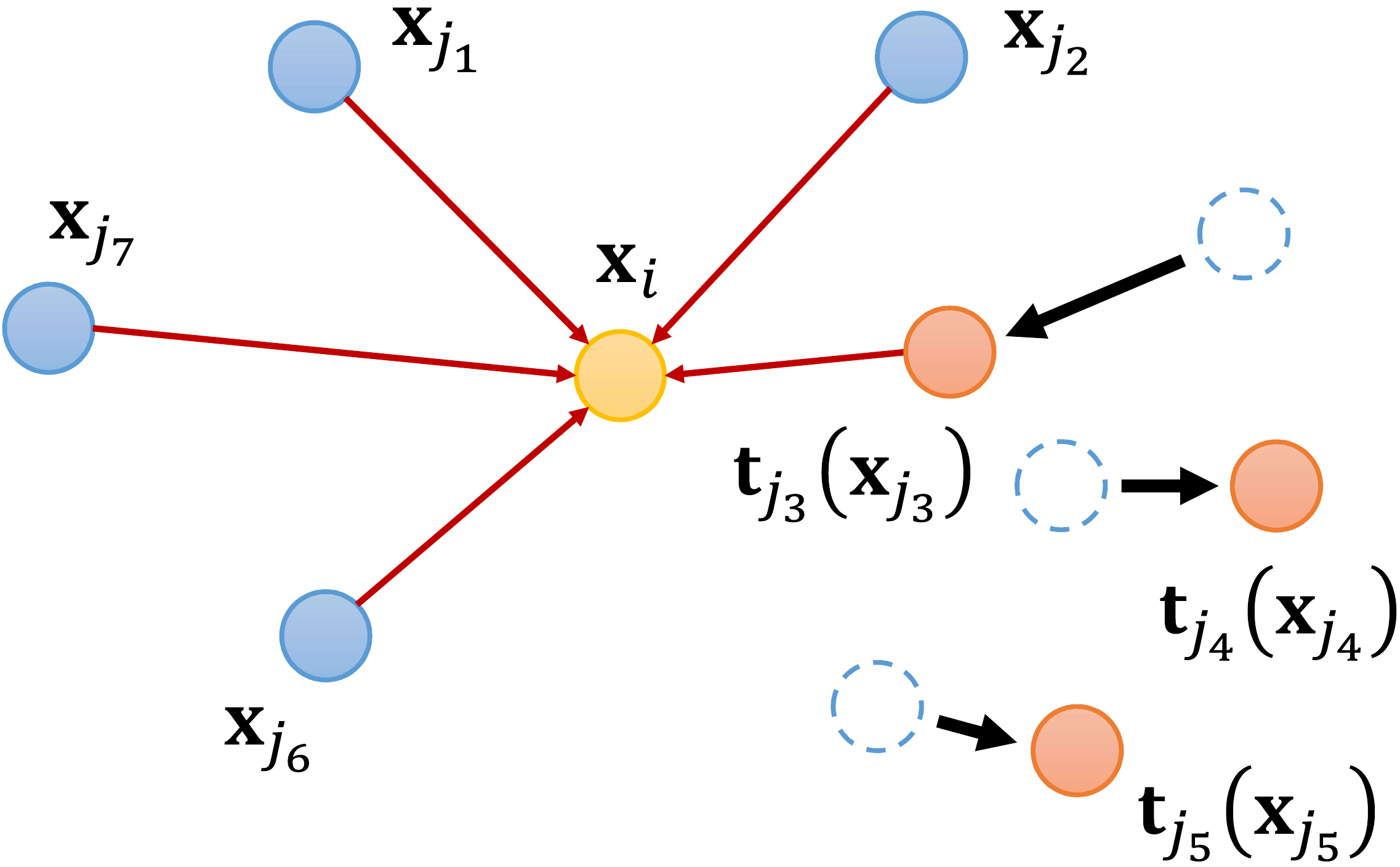}
 	\label{subfig:after_transform}
 	}
    \caption{\textbf{An example of $k$NN graphs before and after node-wise transformations.} We first construct a $k$NN ($k=5$) graph for the yellow node (other connections are omitted). Then we perform node-wise transformations on some blue nodes, which alters the graph structure around the yellow node.}
    \label{fig:node_wise_transform}
    \vspace{-0.2in}
\end{figure}

\vspace{-0.05in}
\subsection{The Formulation}

Given a pair of graph signal and adjacency matrix $(\mathbf{X}, \mathbf{A})$, and a pair of {\em transformed} graph signal and adjacency matrix $(\tilde{\mathbf{X}},\tilde{\mathbf{A}})$ by a node-wise graph transformation $\mathbf{t}$, a function $E(\cdot)$ is \textit{transformation equivariant} if it satisfies
\begin{equation}
    \setlength{\abovedisplayskip}{3pt}
    \setlength{\belowdisplayskip}{3pt}
    E(\tilde{\mathbf{X}}, \tilde{\mathbf{A}}) = E\left(\mathbf{t}(\mathbf{X}), f\left(\mathbf{t}(\mathbf{X})\right)\right) = \rho(\mathbf{t})\left[E(\mathbf{X}, \mathbf{A})\right],
    \label{eq:ter}
\end{equation}
where $\rho(\mathbf{t})$ is a homomorphism of transformation $\mathbf{t}$ in the representation space.

Our goal is to learn a function $E(\cdot)$, which extracts equivariant representations of graph signals $\mathbf{X}$. 
For this purpose, we employ an encoder-decoder network: we learn a graph encoder $E: (\mathbf{X},\mathbf{A}) \mapsto E(\mathbf{X},\mathbf{A})$, which encodes the feature representations of individual nodes from the graph. 
To ensure the transformation equivariance of representations, we train a decoder $D: \left(E(\mathbf{X},\mathbf{A}), E(\tilde{\mathbf{X}},\tilde{\mathbf{A}})\right) \mapsto \hat{\mathbf{t}}$ to estimate the node-wise transformation $\hat{\mathbf{t}}$ from the representations of the original and transformed graph signals. 
Hence, we cast the learning problem of transformation equivariant representations as the joint training of the representation encoder $E$ and the transformation decoder $D$. It has been proved that the learned representations in this way satisfy the generalized transformation equivariance without relying on a linear representation of graph structures \cite{qi2019avt}. 

Further, we sample a subset of nodes $\mathbf S$ following a sampling distribution $\mathcal S_g$ from the original graph signal $\mathbf X$, locally or globally in order to reveal graph structures at various scales. 
Node-wise transformations are then performed on the subset $\mathbf S$ isotropically or anisotropically, as demonstrated in Fig.~\ref{fig:sampling_transform}. 
In order to predict the node-wise transformation $\mathbf{t}$, we choose a loss function $\ell_\mathbf S(\mathbf{t}, \hat{\mathbf{t}})$ that quantifies the distance between $\mathbf{t}$ and its estimate $\hat{\mathbf{t}}$ in terms of their parameters. 
Then the entire network is trained end-to-end by minimizing the loss
\begin{equation}
    \setlength{\abovedisplayskip}{3pt}
    \setlength{\belowdisplayskip}{3pt}
    \min_{E,D} \;
    \underset{\mathbf S \sim \mathcal S_g}{\mathbb E}~\underset{\mathbf X \sim \mathcal X_g}{\underset{\mathbf{t} \sim \mathcal{T}_g}{\mathbb E}}
     ~ \ell_\mathbf S(\mathbf{t}, \hat{\mathbf{t}}),
    \label{eq:loss}
\end{equation}
where the expectation $\mathbb{E}$ is taken over the sampled graph signals and transformations, and the loss is taken over the (locally or globally) sampled subset $\mathbf S$ of nodes in each iteration of training. 

In \eqref{eq:loss}, the node-wise transformation $\hat{\mathbf{t}}$ is estimated from the decoder
\begin{equation}
    \setlength{\abovedisplayskip}{3pt}
    \setlength{\belowdisplayskip}{3pt}
    \hat{\mathbf{t}} = D\left(E(\mathbf{X},\mathbf{A}), E(\tilde{\mathbf{X}},\tilde{\mathbf{A}})\right).
\end{equation}
Thus, we update the parameters in encoder $E$ and decoder $D$ iteratively by backward propagation of the loss.

 
\begin{figure}[t]
    \centering
    \includegraphics[width=8.3cm]{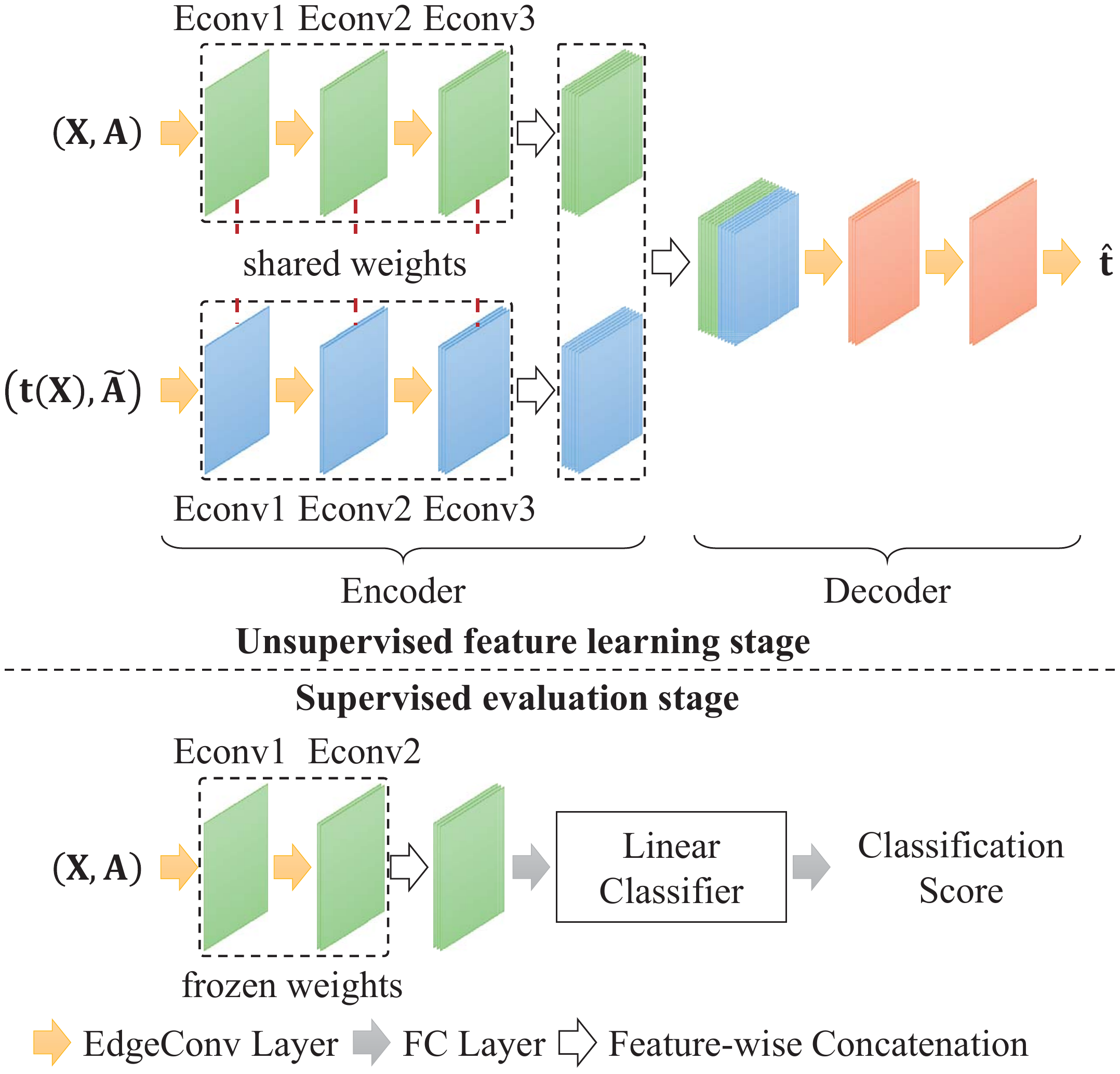}
    \caption{\textbf{The architecture of the proposed GraphTER.} 
    In the unsupervised feature learning stage, the representation encoder and transformation decoder are jointly trained by minimizing (\ref{eq:loss}). In the supervised evaluation stage, the first several blocks of the encoder are fixed with frozen weights and a linear classifier is trained with labeled samples.
    }
    \label{fig:framework}
    \vspace{-0.2in}
\end{figure}

\vspace{-0.05in}
\subsection{The Algorithm}

Given graph signals $\mathbf{X}=\{\mathbf{x}_1,\mathbf{x}_2,...,\mathbf{x}_N\}^{\top}$ over $N$ nodes, in each iteration of training, we \textit{randomly sample} a subset of nodes $\mathbf S$ from the graph, either globally or locally. 
Global sampling refers to random sampling over the entire nodes globally, while local sampling is limited to a local set of nodes in the graph.
Node sampling not only enables us to characterize global and local graph structures at various scales, but also reduces the number of node-wise transformation parameters to estimate for computational efficiency.

Then we draw a node-wise transformation $\mathbf{t}_i$ corresponding to each sample $\mathbf{x}_i$ of nodes in $\mathbf S$, either isotropically or anisotropically. 
Accordingly, the graph $\tilde{\mathbf{A}}$ associated with the transformed graph also transforms equivariantly from the original $\mathbf{A}$ under $\mathbf t$.
Specifically, as illustrated in Fig.~\ref{fig:node_wise_transform}, we construct a $k$NN graph to make use of the connectivity between the nodes, whose matrix representation in $\mathbf{A}$ changes after applying the sampled node-wise transformations. 

To learn the applied node-wise transformations, we design a full graph-convolutional auto-encoder network as illustrated in Fig.~\ref{fig:framework}.
Among various paradigms of GCNNs, we choose EdgeConv \cite{wang2019dynamic} as a basic building block of the auto-encoder network, which efficiently learns node-wise representations by aggregating features along all the edges emanating from each connected node. Below we will explain the representation encoder and the transformation decoder in detail.

\vspace{-0.05in}
\subsubsection{Representation Encoder}

The representation encoder $E$ takes the signals of an original graph $\mathbf{X}$ and the transformed counterparts $\tilde{\mathbf{X}}$ as input, along with their corresponding graphs. 
$E$ encodes node-wise features of $\mathbf{X}$ and $\tilde{\mathbf{X}}$ through a Siamese encoder network with shared weights, where EdgeConv layers are used as basic feature extraction blocks.
As shown in Fig.~\ref{fig:node_wise_transform}, given a non-transformed central node $\mathbf{x}_i$ and its transformed neighbors $\mathbf{t}_j(\mathbf{x}_j)$, the input layer of encoded feature of $\mathbf{x}_i$ is
\begin{equation}
\setlength{\abovedisplayskip}{3pt}
\setlength{\belowdisplayskip}{3pt}
\begin{split}
    E_{\text{in}}(\tilde{\mathbf{X}}, \tilde{\mathbf{A}})_i & = \underset{j \in \mathcal{N}(i)}{\max} \; \tilde{a}_{i,j} \\
    & = \underset{j \in \mathcal{N}(i)}{\max} \; \text{ReLU}(\theta(\mathbf{t}_j(\mathbf{x}_j)-\mathbf{x}_i) + \phi \mathbf{x}_i),
\end{split}
\label{eq:encoding}
\end{equation}
where $\tilde{a}_{i,j}$ denotes the edge feature, {\it i.e.}, edge weight in $\tilde{\mathbf{A}}$. 
$\theta$ and $\phi$ are two weighting parameters,
and $j \in \mathcal{N}(i)$ denotes node $j$ is in the $k$-nearest neighborhood of node $i$. 
Then, multiple layers of regular edge convolutions \cite{wang2019dynamic} are stacked to form the final encoder.

Edge convolution in (\ref{eq:encoding}) over each node essentially aggregates features from neighboring nodes via {\it edge weights} $\tilde{a}_{i,j}$. 
Since the edge information of the underlying graph transforms with the transformations of individual nodes as demonstrated in Fig.~\ref{fig:node_wise_transform}, edge convolution is able to extract higher-level features from the original and transformed edge information.
Also, as features of each node are learned via propagation from transformed and non-transformed nodes isotropically or anisotropically by both local or global sampling, the learned representation is able to capture intrinsic graph structures at multiple scales.   

\vspace{-0.05in}
\subsubsection{Transformation Decoder}

Node-wise features of the original and transformed graphs are then concatenated at each node, which are then fed into the transformation decoder.
The decoder consists of several EdgeConv blocks to aggregate the representations of both the original and transformed graphs to predict the node-wise transformations $\mathbf{t}$. 
Based on the loss in \eqref{eq:loss}, $\mathbf{t}$ is decoded by minimizing the mean squared error (MSE) between the ground truth and estimated transformation parameters at each sampled node. Fig.~\ref{fig:framework} illustrates the architecture of learning the proposed GraphTER in such an auto-encoder structure.

\vspace{-0.1in}
\section{Experiments}
\vspace{-0.05in}
\label{sec:experiments}
In this section, we evaluate the GraphTER model by applying it to graphs of 3D point cloud data on two representative downstream tasks: point cloud classification and segmentation. 
We compare the proposed method with state-of-the-art supervised and unsupervised approaches.

\vspace{-0.05in}
\subsection{Datasets and Experimental Setup}

\textbf{ModelNet40} \cite{wu20153d}. This dataset contains $12,311$ meshed CAD models from $40$ categories, where $9,843$ models are used for training and $2,468$ models are for testing. 
For each model, $1,024$ points are sampled from the original mesh.
We train the unsupervised auto-encoder and the classifier under the training set, and evaluate the classifier under the testing set.

\textbf{ShapeNet part} \cite{yi2016scalable}. This dataset contains $16,881$ 3D point clouds from $16$ object categories, annotated with $50$ parts. 
Each 3D point cloud contains $2,048$ points, most of which are labeled with fewer than six parts.
We employ $12,137$ models for training the auto-encoder and the classifier, and $2,874$ models for testing.

We treat points in each point cloud as nodes in a graph, and the $(x,y,z)$ coordinates of points as graph signals. 
A $k$NN graph is then constructed on the graph signals to guide graph convolution.

Next, we introduce our node-wise graph signal transformation. In experiments, we sample a portion of nodes with a sampling rate $r$ from the entire graph to perform node-wise transformations, including 
1) \textbf{Global sampling:} randomly sample $r\%$ of points from all the points in a 3D point cloud;
2) \textbf{Local sampling:} randomly choose a point and search its $k$ nearest neighbors in terms of Euclidean distance, forming a local set of $r\%$ of points.

Then, we apply three types of node-wise transformations to the coordinates of point clouds, including 
1) \textbf{Translation:} randomly translate each of three coordinates of a point by three parameters in the range $[-0.2, 0.2]$;
2) \textbf{Rotation:} randomly rotate each point with three rotation parameters all in the range $[\ang{-5},\ang{5}]$;
3) \textbf{Shearing}: randomly shear the $x$-, $y$-, $z$-coordinates of each point with the six parameters of a shearing matrix in the range $[-0.2, 0.2]$.
We consider two strategies to transform the sampled nodes: \textbf{Isotropically} or \textbf{Anisotropically}, which applies transformations with node-invariant or node-variant parameters.

\begin{table}[t]
\centering
\small
\caption{Classification accuracy (\%) on ModelNet40 dataset.}
\label{tab:classification}
\begin{tabular}{rcccc}
\hline
\multicolumn{1}{c}{\textbf{Method}} & \textbf{Year} & \textbf{Unsupervised} & \textbf{Accuracy} \\ \hline
3D ShapeNets \cite{wu20153d} & 2015 & No & 84.7 \\
VoxNet \cite{maturana2015voxnet} & 2015 & No & 85.9 \\
PointNet \cite{qi2017pointnet} & 2017 & No & 89.2 \\
PointNet++ \cite{qi2017pointnet++} & 2017 & No & 90.7 \\
KD-Net \cite{klokov2017escape} & 2017 & No & 90.6 \\
PointCNN \cite{li2018pointcnn} & 2018 & No & 92.2 \\
PCNN \cite{atzmon2018point} & 2018 & No & 92.3 \\
DGCNN \cite{wang2019dynamic} & 2019 & No & 92.9 \\
RS-CNN \cite{liu2019relation} & 2019 & No & 93.6 \\ \hline
T-L Network \cite{girdhar2016learning} & 2016 & Yes & 74.4 \\
VConv-DAE \cite{sharma2016vconv} & 2016 & Yes & 75.5 \\
3D-GAN \cite{wu2016learning} & 2016 & Yes & 83.3 \\
LGAN \cite{achlioptas2018learning} & 2018 & Yes & 85.7 \\
FoldingNet \cite{yang2018foldingnet} & 2018 & Yes & 88.4 \\
MAP-VAE \cite{han2019multi} & 2019 & Yes & 90.2 \\
L2G-AE \cite{liu2019l2g} & 2019 & Yes & 90.6 \\ \hline
GraphTER & & Yes & \textbf{92.0} \\ \hline 
\end{tabular}
\vspace{-0.2in}
\end{table}

\vspace{-0.05in}
\subsection{Point Cloud Classification}

First, we evaluate the GraphTER model on the ModelNet40 \cite{wu20153d} dataset for point cloud classification.

\vspace{-0.1in}
\subsubsection{Implementation Details}

In this task, the auto-encoder network is trained via the SGD optimizer with a batch size of $32$. 
The momentum and weight decay rate are set to $0.9$ and $10^{-4}$, respectively. 
The initial learning rate is $0.1$, and then decayed using a cosine annealing schedule \cite{loshchilov2016sgdr} for $512$ training epochs. 
We adopt the cross entropy loss to train the classifier.

We deploy eight EdgeConv layers as the encoder, and the number $k$ of nearest neighbors is set to $20$ for all EdgeConv layers. 
Similar to \cite{wang2019dynamic}, we use shortcut connections for the first five layers to extract multi-scale features, where we concatenate features from these layers to acquire a $1,024$-dimensional node-wise feature vector. 
After the encoder, we employ three consecutive EdgeConv layers as the decoder -- the output feature representations of the Siamese encoder first go through a channel-wise concatenation, which are then fed into the decoder to estimate node-wise transformations.
The batch normalization layer and LeakyReLU activation function with a negative slope of $0.2$ is employed after each convolutional layer.

During the training procedure of the classifier, the first five EdgeConv layers in the encoder are used to represent input cloud data by node-wise concatenating their output features with the weights frozen. After the five EdgeConv layers, we apply three fully-connected layers node-wise to the aggregated features. Then, global max pooling and average pooling are deployed to acquire the global features, after which three fully-connected layers are used to map the global features to the classification scores. Dropout with a rate of $0.5$ is adopted in the last two fully-connected layers.

\vspace{-0.1in}
\subsubsection{Experimental Results}
Tab.~\ref{tab:classification} shows the results for 3D point cloud classification, where the proposed model applies isotropic node-wise shearing transformation with a global sampling rate of $r=25\%$.
We compare with two classes of methods: unsupervised approaches and supervised approaches. 
The GraphTER model achieves 92.0\% of classification accuracy on the ModelNet40 dataset, which outperforms the state-of-the-art unsupervised methods. 
In particular, most of the compared unsupervised models combine the ideas of both GAN and AED, and map 3D point clouds to unsupervised representations by auto-encoding data, such as FoldingNet \cite{yang2018foldingnet}, MAP-VAE \cite{han2019multi} and L2G-AE \cite{liu2019l2g}. 
Results show that the GraphTER model achieves significant improvement over these methods, showing the superiority of the proposed node-wise AET over both the GAN and AED paradigms. 

Moreover, the unsupervised GraphTER model also achieves comparable performance with the state-of-the-art fully supervised results.
This significantly closes the gap between unsupervised approaches and the fully supervised counterparts in literature. 

\vspace{-0.1in}
\subsubsection{Ablation Studies}

Further, we conduct ablation studies under various experimental settings of sampling and transformation strategies on the ModelNet40 dataset.

First, we analyze the effectiveness of different node-wise transformations under global or local sampling.
Tab.~\ref{tab:diff_transform} presents the classification accuracy with three types of node-wise transformation methods.
We see that the shearing transformation achieves the best performance,  improving by 1.05\% on average over translation, and 0.52\% over rotation.
This shows that the proposed GraphTER model is able to learn better feature representations under more complex transformations.

Moreover, we see that the proposed model achieves an accuracy of 90.70\% on average under global sampling, which outperforms local sampling by 0.46\%.
This is because global sampling better captures the global structure of graphs, which is crucial in such a graph-level task of classifying 3D point clouds.
Meanwhile, under the two sampling strategies, the classification accuracy from isotropic transformations is higher than that from the anisotropic one.
The reason lies in the intrinsic difficulty of training the transformation decoder with increased complexity of more parameters when applying anisotropic transformations. 

\begin{table}[t]
\centering
\small
\caption{Unsupervised classification accuracy (\%) on ModelNet40 dataset with different sampling and transformation strategies.}
\label{tab:diff_transform}
\begin{tabular}{c|cc|cc|c}
\hline
\multirow{2}{*}{} & \multicolumn{2}{c|}{Global Sampling} & \multicolumn{2}{c|}{Local Sampling} & \multirow{2}{*}{Mean} \\ \cline{2-5}
 & Iso. & Aniso. & Iso. & Aniso. & \\ \hline
Translation & 90.15 & 90.15 & 89.91 & 89.55 & 89.94 \\
Rotation & 91.29 & 90.24 & 90.48 & 89.87 & 90.47 \\
Shearing & \textbf{92.02} & \textbf{90.32} & \textbf{91.65} & \textbf{89.99} & \textbf{90.99} \\ \hline
\multirow{2}{*}{Mean} & \textbf{91.15} & 90.24 & \textbf{90.68} & 89.80 & \\ \cline{2-5}
 & \multicolumn{2}{c|}{\textbf{90.70}} & \multicolumn{2}{c|}{90.24} &  \\ \hline
\end{tabular}
\vspace{-0.1in}
\end{table}

Moreover, we evaluate the effectiveness of different sampling rates $r$ under the translations as reported in Tab.~\ref{tab:sample_rate}.
The classification accuracies under various sampling rates are almost the same, and the result under $r=25\%$ is comparable to that under $r=100\%$. 
This shows that the performance of the proposed model is insensitive to the variation of sampling rates, {\it i.e.}, applying node-wise transformations to a small number of nodes in the graph is sufficient to learn intrinsic graph structures.


\begin{table}[t]
\centering
\small
\caption{Unsupervised classification accuracy (\%) on ModelNet40 dataset applying translation at different node sampling rates.}
\label{tab:sample_rate}
\begin{tabular}{c|cc|cc|c}
\hline
\multirow{2}{*}{\begin{tabular}[c]{@{}c@{}}Sampling\\ Rate\end{tabular}} & \multicolumn{2}{c}{Global Sampling} & \multicolumn{2}{c|}{Local Sampling} & \multirow{2}{*}{Mean} \\ \cline{2-5}
 & Iso. & Aniso. & Iso. & Aniso. & \\ \hline
25\% & 90.15 & 90.15 & 89.91 & 89.55 & 89.94 \\
50\% & 90.03 & 89.63 & 89.95 & 89.47 & 89.77 \\
75\% & 91.00 & 89.67 & 91.41 & 89.75 & 90.46 \\
100\% & 89.67 & 89.99 & 89.67 & 89.99 & 89.83 \\ \hline
\end{tabular}
\vspace{-0.1in}
\end{table}

\begin{table*}[t]
\centering
\scriptsize
\caption{Part segmentation results on ShapeNet part dataset. Metric is mIoU(\%) on points.}
\label{tab:segmentation}
\begin{tabular}{r|p{0.65cm}<{\centering}|p{0.5cm}<{\centering}|p{0.4cm}<{\centering}p{0.4cm}<{\centering}p{0.4cm}<{\centering}p{0.4cm}<{\centering}p{0.4cm}<{\centering}p{0.4cm}<{\centering}p{0.4cm}<{\centering}p{0.4cm}<{\centering}p{0.4cm}<{\centering}p{0.4cm}<{\centering}p{0.4cm}<{\centering}p{0.4cm}<{\centering}p{0.4cm}<{\centering}p{0.4cm}<{\centering}p{0.4cm}<{\centering}p{0.4cm}<{\centering}}
\hline
 & Unsup. & Mean & Aero & Bag & Cap & Car & Chair & \begin{tabular}[c]{@{}c@{}}Ear\\ Phone\end{tabular} & Guitar & Knife & Lamp & Laptop & Motor & Mug & Pistol & Rocket & \begin{tabular}[c]{@{}c@{}}Skate\\ Board\end{tabular} & Table \\ \hline
Samples & & & 2690 & 76 & 55 & 898 & 3758 & 69 & 787 & 392 & 1547 & 451 & 202 & 184 & 283 & 66 & 152 & 5271 \\ \hline
PointNet \cite{qi2017pointnet} & No & 83.7 & 83.4 & 78.7 & 82.5 & 74.9 & 89.6 & 73.0 & 91.5 & 85.9 & 80.8 & 95.3 & 65.2 & 93.0 & 81.2 & 57.9 & 72.8 & 80.6 \\
PointNet++ \cite{qi2017pointnet++} & No & 85.1 & 82.4 & 79.0 & 87.7 & 77.3 & 90.8 & 71.8 & 91.0 & 85.9 & 83.7 & 95.3 & 71.6 & 94.1 & 81.3 & 58.7 & 76.4 & 82.6 \\
KD-Net \cite{klokov2017escape} & No & 82.3 & 80.1 & 74.6 & 74.3 & 70.3 & 88.6 & 73.5 & 90.2 & 87.2 & 81.0 & 94.9 & 57.4 & 86.7 & 78.1 & 51.8 & 69.9 & 80.3 \\
PCNN \cite{atzmon2018point} & No & 85.1 & 82.4 & 80.1 & 85.5 & 79.5 & 90.8 & 73.2 & 91.3 & 86.0 & 85.0 & 95.7 & 73.2 & 94.8 & 83.3 & 51.0 & 75.0 & 81.8 \\
PointCNN \cite{li2018pointcnn} & No & 86.1 & 84.1 & 86.5 & 86.0 & 80.8 & 90.6 & 79.7 & 92.3 & 88.4 & 85.3 & 96.1 & 77.2 & 95.3 & 84.2 & 64.2 & 80.0 & 83.0 \\
DGCNN \cite{wang2019dynamic} & No & 85.2 & 84.0 & 83.4 & 86.7 & 77.8 & 90.6 & 74.7 & 91.2 & 87.5 & 82.8 & 95.7 & 66.3 & 94.9 & 81.1 & 63.5 & 74.5 & 82.6 \\
RS-CNN \cite{liu2019relation} & No & 86.2 & 83.5 & 84.8 & 88.8 & 79.6 & 91.2 & 81.1 & 91.6 & 88.4 & 86.0 & 96.0 & 73.7 & 94.1 & 83.4 & 60.5 & 77.7 & 83.6 \\ \hline
LGAN \cite{achlioptas2018learning} & Yes & 57.0 & 54.1 & 48.7 & 62.6 & 43.2 & 68.4 & 58.3 & 74.3 & 68.4 & 53.4 & 82.6 & 18.6 & 75.1 & 54.7 & 37.2 & 46.7 & 66.4 \\
MAP-VAE \cite{han2019multi} & Yes & 68.0 & 62.7 & 67.1 & 73.0 & 58.5 & 77.1 & 67.3 & 84.8 & 77.1 & 60.9 & 90.8 & 35.8 & 87.7 & 64.2 & 45.0 & 60.4 & 74.8 \\ \hline
GraphTER & Yes & \textbf{81.9} & \textbf{81.7} & \textbf{68.1} & \textbf{83.7} & \textbf{74.6} & \textbf{88.1} & \textbf{68.9} & \textbf{90.6} & \textbf{86.6} & \textbf{80.0} & \textbf{95.6} & \textbf{56.3} & \textbf{90.0} & \textbf{80.8} & \textbf{55.2} & \textbf{70.7} & \textbf{79.1} \\ \hline
\end{tabular}
\vspace{-0.2in}
\end{table*}

\subsection{Point Cloud Segmentation}

We also apply the GraphTER model to 3D point cloud part segmentation on ShapeNet part dataset \cite{yi2016scalable}.

\vspace{-0.2in}
\subsubsection{Implementation Details}

We also use SGD optimizer to train the auto-encoding transformation network. 
The hyper-parameters are the same as in 3D point cloud classification except that we train for $256$ epochs. 
We adopt the negative log likelihood loss to train the node-wise classifier for segmenting each point in the clouds.

The auto-encoding architecture is similar to that of the classification task, where we employ five EdgeConv layers as the encoder.
However, the first two EdgeConv blocks consist of two MLP layers with the number of neurons \{64, 64\} in each layer.
We use shortcut connections to concatenate features from the first four layers to  a $512$-dimensional node-wise feature vector.

As for the node-wise classifier, we deploy the same architecture as in \cite{wang2019dynamic}. The output features from the encoder are concatenated node-wise with globally max-pooled features, followed by four fully-connected layers to classify each node. 
During the training procedure, the weights of the first four EdgeConv blocks in the encoder are kept frozen.

\vspace{-0.2in}
\subsubsection{Experimental Results}

We adopt the Intersection-over-Union (IoU) metric to evaluate the performance.
We follow the same evaluation protocol as in the PointNet \cite{qi2017pointnet}: the IoU of a shape is computed by averaging the IoUs of different parts occurring in that shape, and the IoU of a category is obtained by averaging the IoUs of all the shapes belonging to that category.
The mean IoU (mIoU) is finally calculated by averaging the IoUs of all the test shapes.

We also compare the proposed model with unsupervised approaches and supervised approaches in this task, as listed in Tab.~\ref{tab:segmentation}. 
We achieve a mIoU of $81.9\%$, which significantly outperforms the state-of-the-art unsupervised method MAP-VAE \cite{han2019multi} by $13.9\%$.

Moreover, the unsupervised GraphTER model also achieves the comparable performance to the state-of-the-art fully supervised approaches, greatly pushing closer towards the upper bound set by the fully supervised counterparts.

\vspace{-0.2in}
\subsubsection{Visualization Results}

\begin{figure}[t]
    \centering
    \subfigure[Ground-truth]{
    \includegraphics[width=0.22\columnwidth]{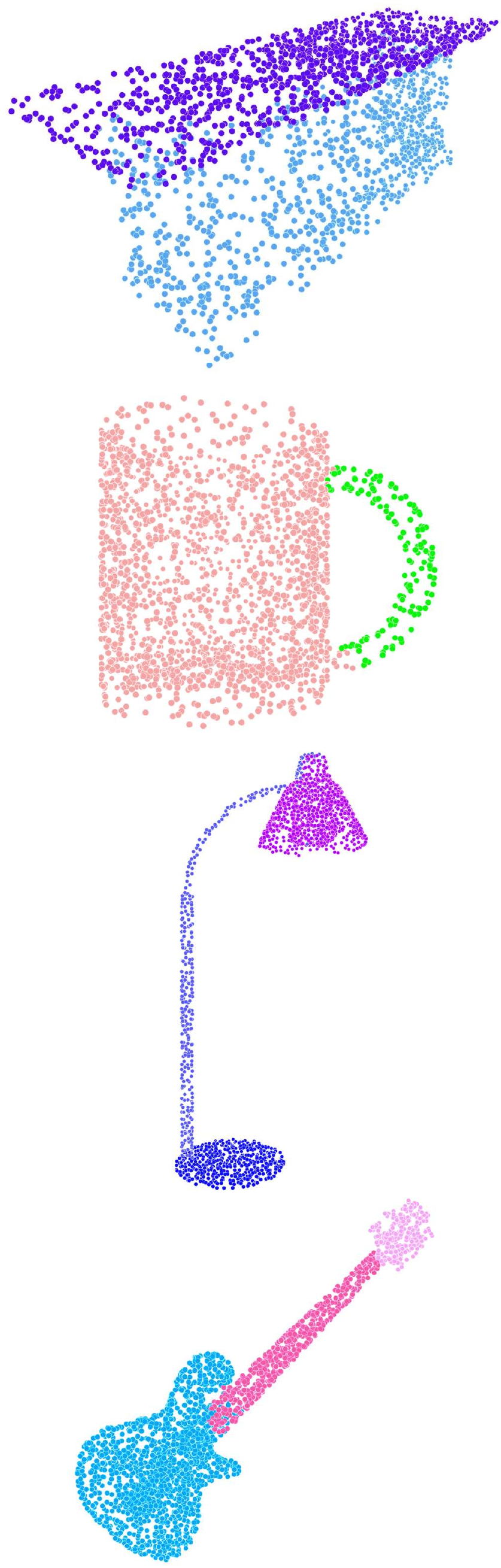}
    }
 	\subfigure[DGCNN]{
 	\includegraphics[width=0.22\columnwidth]{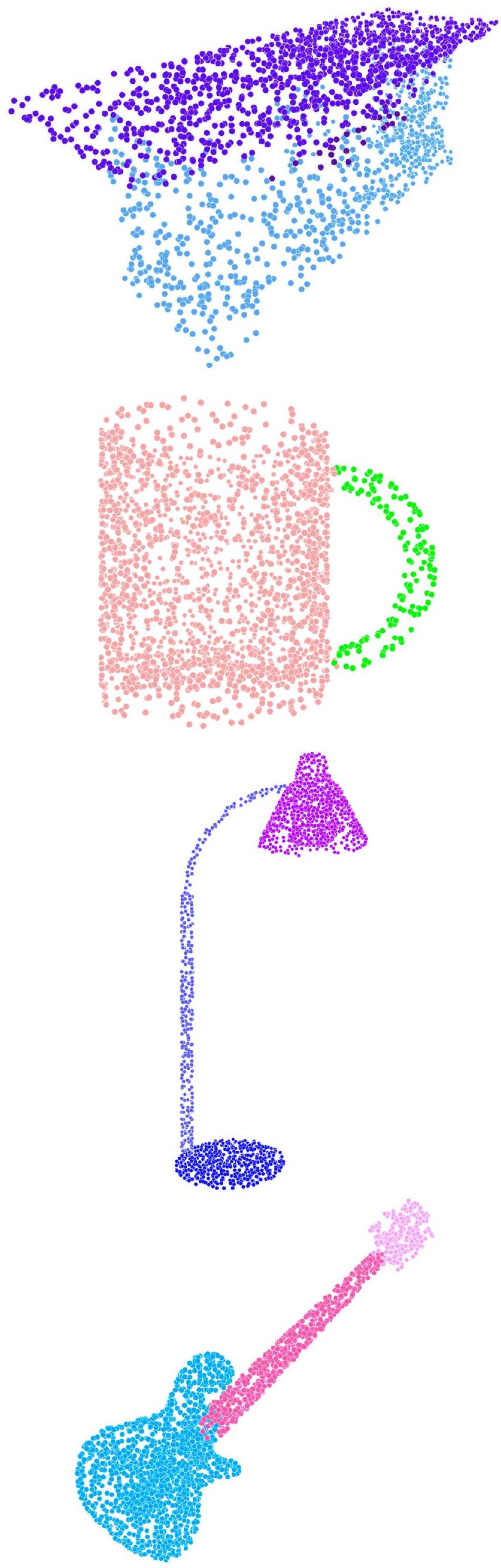}
 	}
 	\subfigure[RS-CNN]{
 	\includegraphics[width=0.22\columnwidth]{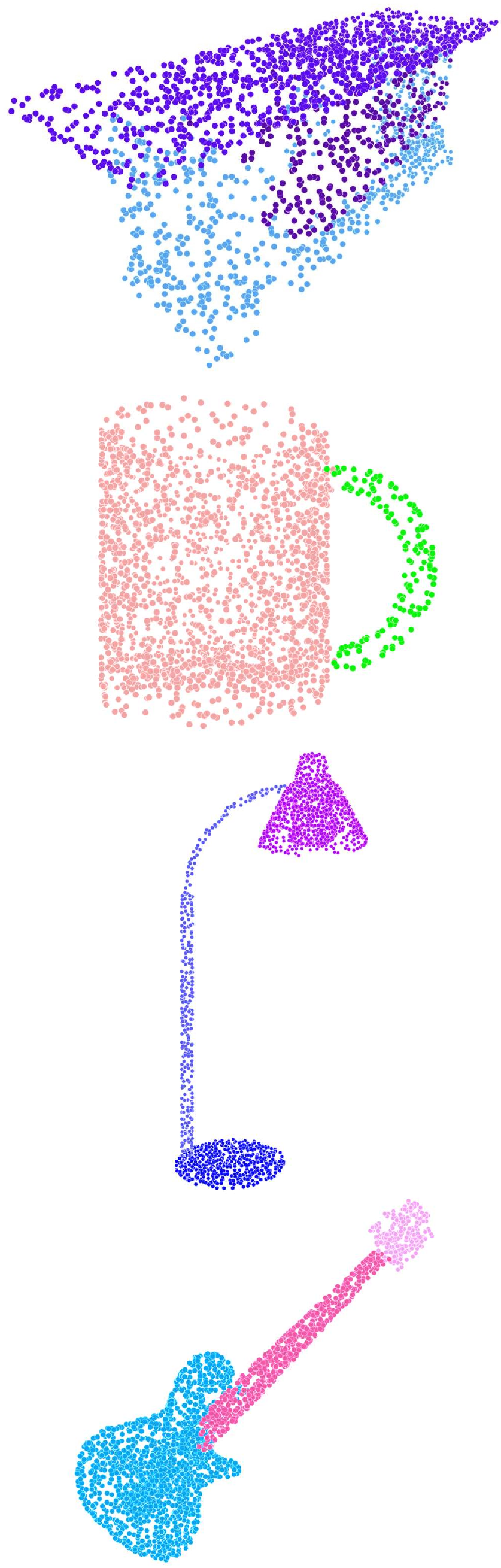}
 	}
 	\subfigure[GraphTER]{
 	\includegraphics[width=0.22\columnwidth]{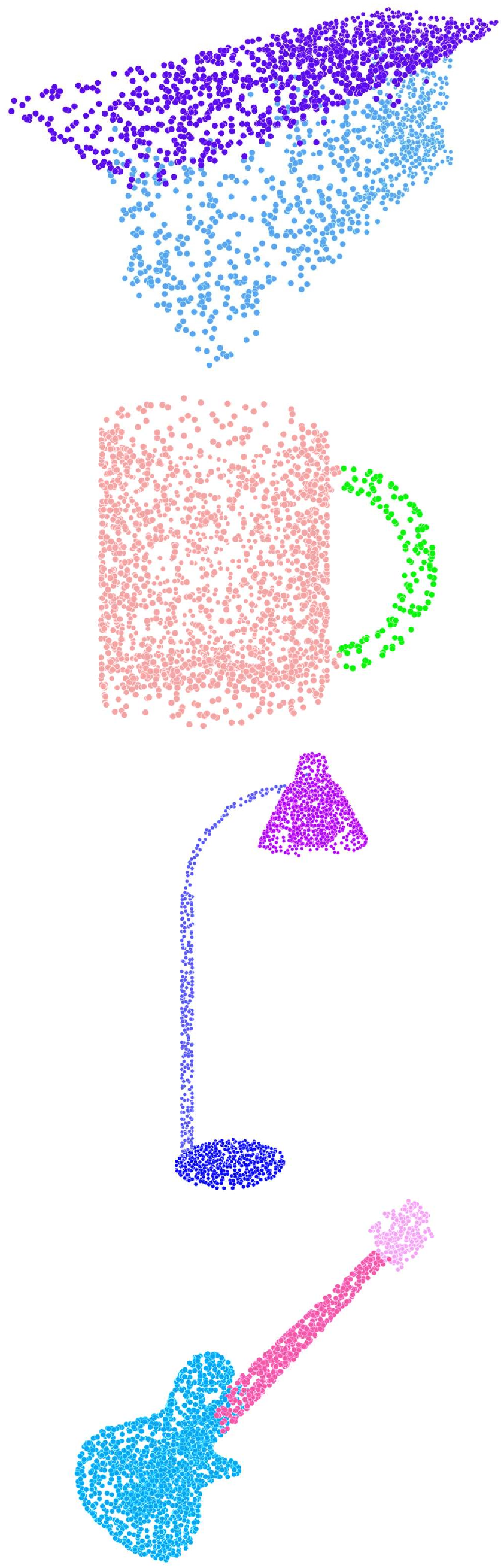}
 	}
    \caption{\textbf{Visual comparison of point cloud part segmentation with supervised methods. } 
    Our unsupervised GraphTER learning achieves comparable results with the state-of-the art fully supervised approaches.
    }
    \label{fig:sup_seg_results}
    \vspace{-0.1in}
\end{figure}

\begin{figure}[t]
    \centering
 	\subfigure[MAP-VAE]{
 	\includegraphics[width=0.95\columnwidth]{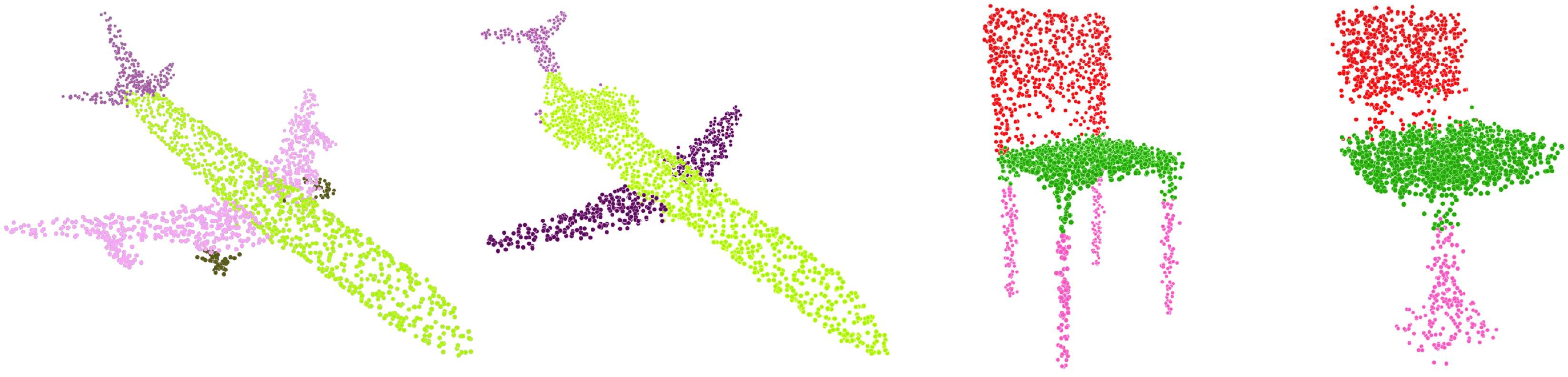}
 	}
 	\subfigure[GraphTER]{
 	\includegraphics[width=0.95\columnwidth]{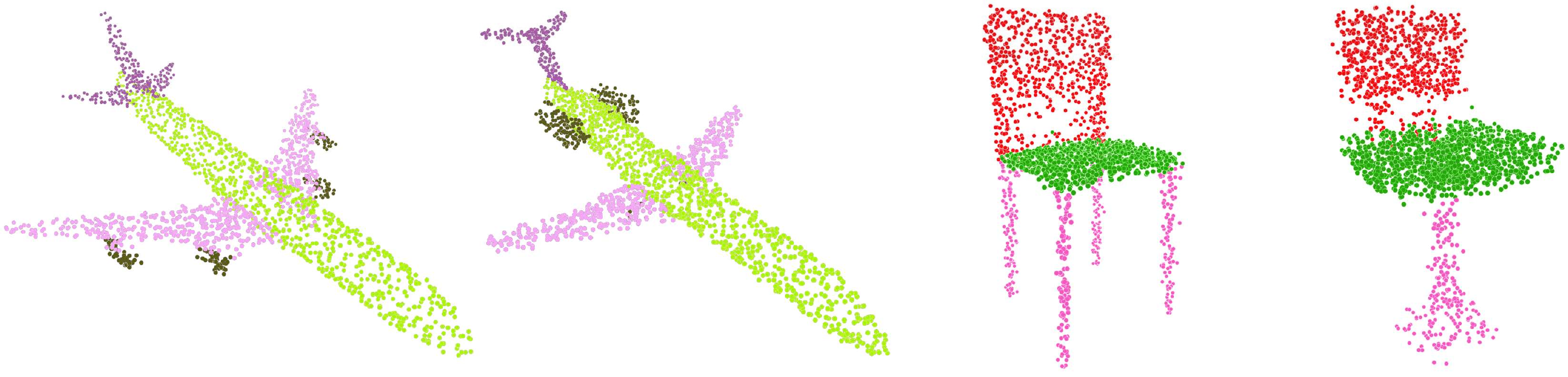}
 	}
    \caption{\textbf{Visual comparison of point cloud part segmentation with the state-of-the-art unsupervised method MAP-VAE. } 
    We achieve more accurate segmentation even in tiny parts and transition regions.
    }
    \label{fig:unsup_seg_results}
    \vspace{-0.1in}
\end{figure}

Fig.~\ref{fig:sup_seg_results} visualizes the results of the proposed unsupervised model and two state-of-the-art fully supervised methods: DGCNN \cite{wang2019dynamic} and RS-CNN \cite{liu2019relation}.
The proposed model produces better segmentation on the ``table" model in the first row, and achieves comparable results on the other models.
Further, we qualitatively compare the proposed method with the state-of-the-art unsupervised method MAP-VAE \cite{han2019multi}, as illustrate in Fig.~\ref{fig:unsup_seg_results}. 
The proposed model leads to more accurate segmentation results than MAP-VAE, {\it e.g.}, the engines of planes and the legs of chairs.

\vspace{-0.1in}
\subsubsection{Ablation Studies}

\begin{table}[t]
\centering
\small
\caption{Unsupervised segmentation results on ShapeNet part dataset with different transformation strategies. Metric is mIoU (\%) on points.}
\label{tab:seg_ablation}
\begin{tabular}{c|cc|cc|c}
\hline
\multirow{2}{*}{} & \multicolumn{2}{c}{Global Sampling} & \multicolumn{2}{c|}{Local Sampling} & \multirow{2}{*}{Mean} \\ \cline{2-5}
 & Iso. & Aniso. & Iso. & Aniso. & \\ \hline
Translation & 79.83 & 79.88 & 80.05 & 79.85 & 79.90 \\
Rotation & 80.20 & \textbf{80.29} & 80.87 & 80.02 & 80.35 \\
Shearing & \textbf{81.88} & 80.28 & \textbf{81.89} & \textbf{80.48} & \textbf{81.13} \\ \hline
\multirow{2}{*}{Mean} & \textbf{80.64} & 80.15 & \textbf{80.94} & 80.12 & \multirow{2}{*}{\textbf{}} \\ \cline{2-5}
 & \multicolumn{2}{c|}{80.39} & \multicolumn{2}{c|}{\textbf{80.53}} & \\ \hline
\end{tabular}
\vspace{-0.2in}
\end{table}

Similar to the classification task, we analyze the effectiveness of different node-wise transformations under global or local sampling, as presented in Tab.~\ref{tab:seg_ablation}.
The proposed model achieves the best performance under the shearing transformation, improving by 1.23\% on average over translation, and 0.78\% over rotation, which demonstrates the benefits of GraphTER learning under complex transformations.

Further, the proposed model achieves a mIoU of 80.53\% on average under local sampling, which outperforms global sampling by 0.14\%.
This is because local sampling of nodes captures the local structure of graphs better, which is crucial in node-level 3D point cloud segmentation task.

\section{Conclusion}
\vspace{-0.05in}
\label{sec:conclusion}
In this paper, we propose a novel  paradigm of learning graph transformation equivariant representation (GraphTER) via auto-encoding node-wise transformations in an unsupervised fashion. 
We allow it to sample different groups of nodes from a graph globally or locally and then perform node-wise transformations isotropically or anisotropically, which enables it to characterize morphable structures of graphs at various scales. 
By decoding these node-wise transformations, GraphTER enforces the encoder to learn intrinsic representations that contain sufficient information about structures under applied transformations.
We apply the GraphTER model to classification and segmentation of graphs of 3D point cloud data, and experimental results demonstrate the superiority of GraphTER over the state-of-the-art unsupervised approaches, significantly closing the gap with the fully supervised counterparts. 
We will apply the general GraphTER model to more applications as future works, such as node classification of citation networks.  

\clearpage

{\small
\bibliographystyle{ieee}

}

\end{document}